%% file: main.tex
\documentclass[runningheads]{llncs}

\usepackage{microtype}
\usepackage{graphicx}
\usepackage{subfigure}
\usepackage{booktabs}
\usepackage{makecell}

\usepackage{algorithm}
\usepackage{algorithmic}

\usepackage{nicefrac}
\usepackage{bm}
\usepackage{booktabs}
\usepackage{amsfonts}
\usepackage{bbm}
\usepackage{mleftright}
\usepackage{dsfont}
\usepackage{placeins}
\usepackage{amscd}
\usepackage{amssymb}
\usepackage{bbold}

\usepackage{hyperref}

\usepackage{booktabs}
\usepackage[svgnames,table]{xcolor}
\usepackage{pifont}

\usepackage{multicol}
\usepackage{multirow}

\usepackage{pdfpages}

\input{mathcommands.tex}

\usepackage{xcolor}

\begin{document}
\title{FedSoup: Improving Generalization and Personalization in Federated Learning via Selective Model Interpolation}

\titlerunning{FedSoup: Federated Model Soups}

\author{Minghui Chen\inst{1}
\and
Meirui Jiang\inst{2} 
\and
Qi Dou
\inst{2}
\and
Zehua Wang\inst{1} 
\and Xiaoxiao Li\inst{1}
}
\authorrunning{M. Chen et al.}

\institute{
Department of Electrical and Computer Engineering,\\The University of British Columbia, Vancouver, Canada
\and 
Department of Computer Science and Engineering, \\ The Chinese University of Hong Kong, Hong Kong}

\maketitle              %
\begin{abstract}
Cross-silo federated learning (FL) enables the development of machine learning models on datasets distributed across data centers such as hospitals and clinical research laboratories.
However, recent research has found that current FL algorithms face a trade-off between local and global performance when confronted with distribution shifts. Specifically, personalized FL methods have a tendency to overfit to local data, leading to a sharp valley in the local model and inhibiting its ability to generalize to out-of-distribution data. 
In this paper, we propose a novel federated model soup method (\textit{i.e.}, selective interpolation of model parameters) to optimize the trade-off between local and global performance.
Specifically, during the federated training phase, each client maintains its own global model pool by monitoring the performance of the interpolated model between the local and global models. This allows us to alleviate overfitting and seek flat minima, which can significantly improve the model's generalization performance.
We evaluate our method on retinal and pathological image classification tasks, and our proposed method achieves significant improvements for out-of-distribution generalization. Our code is available at \href{https://github.com/ubc-tea/FedSoup}{https://github.com/ubc-tea/FedSoup}. \let\thefootnote\relax\footnote{This work is supported in part by the Natural Sciences and Engineering Research Council of Canada (NSERC), Public Safety Canada, Compute Canada and National Natural Science Foundation of China (Project No. 62201485).\\ Corresponding author: Xiaoxiao Li (\email{xiaoxiao.li@ece.ubc.ca})} 

\end{abstract}

\input{sec_1_introduction}

\input{sec_2_method}

\input{sec_3_experiments}

\input{sec_4_conclusion}

\bibliographystyle{splncs04}
\bibliography{main}



\section*{Supplementary material}

# FedSoup: Improving Generalization and Personalization in Federated Learning via Selective Model Interpolation
## (Supplementary Material)

**Table 1:** Test-Time Adaptation (TTA) Global Performance. **Purpose**: Using TTA, a testing stage fine-tuning method, is aimed at comprehensively demonstrating the generalization ability of our proposed method. **Setup**: The experiment follows the settings used in Section 3.2 Table 1 of the main text and we apply Tent [1] (a SOTA TTA method) to baselines and our method during test time. **Results**: As expected, the TTA strategy can improve generalization. It is notable that our proposed method, FedSoup, consistently outperforms other FL methods in terms of accuracy and AUC even after employing the TTA method.

| Method | Pathology | | | | Retinal Fundus | | | |
|---|---|---|---|---|---|---|---|---|
| | Global Perf. | | TTA Global Perf. | | Global Perf. | | TTA Global Perf. | |
| | Accuracy ↑ | AUC ↑ | Accuracy ↑ | AUC ↑ | Accuracy ↑ | AUC ↑ | Accuracy ↑ | AUC ↑ |
| FedAvg | $70.18_{(1.25)}$ | $78.74_{(1.31)}$ | $80.55_{(1.05)}$ | $81.75_{(1.59)}$ | $73.27_{(3.97)}$ | $81.02_{(4.57)}$ | $85.52_{(1.05)}$ | $91.21_{(1.52)}$ |
| FedProx | $67.42_{(1.32)}$ | $77.18_{(1.32)}$ | $79.26_{(0.77)}$ | $79.96_{(0.42)}$ | $70.46_{(2.36)}$ | $76.84_{(2.93)}$ | $84.23_{(1.63)}$ | $91.47_{(1.35)}$ |
| MOON | $70.61_{(1.54)}$ | $79.27_{(1.43)}$ | $84.46_{(1.57)}$ | $88.32_{(1.46)}$ | $77.49_{(3.15)}$ | $84.89_{(2.70)}$ | $86.57_{(1.67)}$ | $93.08_{(0.88)}$ |
| FedBN | $65.16_{(0.69)}$ | $71.61_{(0.97)}$ | $79.39_{(1.14)}$ | $80.87_{(1.87)}$ | $75.11_{(0.52)}$ | $82.70_{(1.01)}$ | $79.15_{(1.29)}$ | $87.46_{(1.17)}$ |
| FedFomo | $61.00_{(0.59)}$ | $61.69_{(0.78)}$ | $80.47_{(1.48)}$ | $80.64_{(1.53)}$ | $59.90_{(2.00)}$ | $63.82_{(1.14)}$ | $61.67_{(2.38)}$ | $65.24_{(1.85)}$ |
| FedRep | $66.87_{(0.60)}$ | $72.34_{(0.75)}$ | $81.64_{(1.14)}$ | $81.23_{(1.46)}$ | $71.81_{(1.79)}$ | $79.09_{(2.44)}$ | $76.40_{(1.18)}$ | $85.20_{(1.51)}$ |
| FedBABU | $69.56_{(1.40)}$ | $77.26_{(1.48)}$ | $82.95_{(0.96)}$ | $84.42_{(0.74)}$ | $77.65_{(0.24)}$ | $85.09_{(0.21)}$ | $86.75_{(1.04)}$ | $93.91_{(0.79)}$ |
| FedSoup | $\mathbf{72.87}_{(1.35)}$ | $\mathbf{81.45}_{(1.40)}$ | $\mathbf{85.36}_{(0.86)}$ | $\mathbf{88.86}_{(1.07)}$ | $\mathbf{78.64}_{(0.90)}$ | $\mathbf{86.24}_{(0.86)}$ | $\mathbf{87.91}_{(1.24)}$ | $\mathbf{94.66}_{(0.65)}$ |

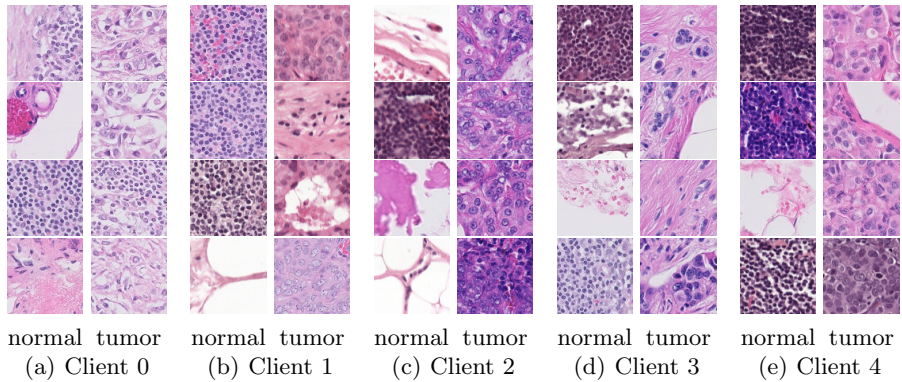

(a) Client 0  (b) Client 1  (c) Client 2  (d) Client 3  (e) Client 4

**Fig. 1:** Camelyon17 pathology dataset visualization.

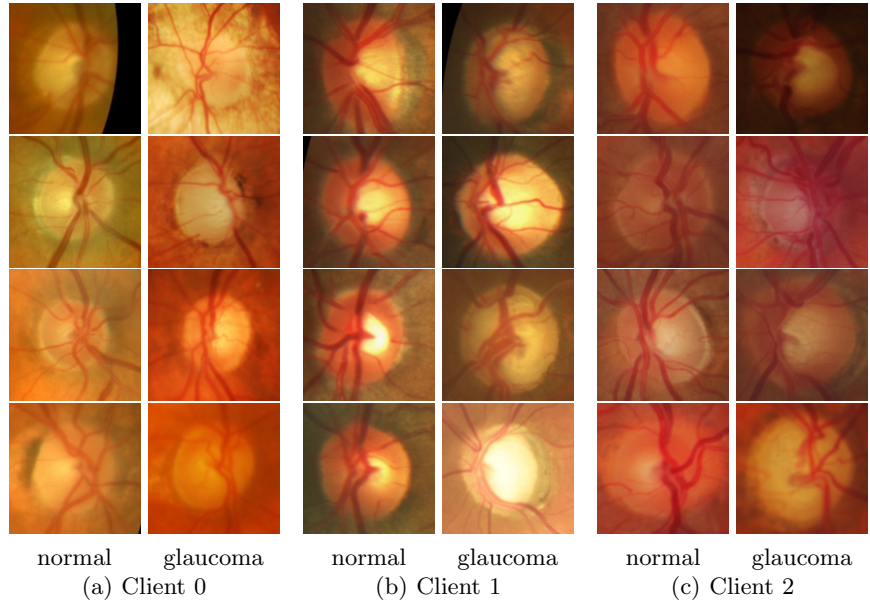

| | normal | glaucoma | normal | glaucoma | normal | glaucoma |
|---|---|---|---|---|---|---|
| | (a) Client 0 | | (b) Client 1 | | (c) Client 2 | |

**Fig. 2:** Retina fundus dataset visualization.

**Table 2:** Additional Results on Unseen Domain Generalization (DG). **Setup**: The experiment follows the settings used in Section 3.3 Figure 2(c) of the main text. **Results**: FedSoup method outperforms other SOTA PFL methods on the pathology dataset and exhibits greater stability with lower variance of performance on unseen domain data. The average performance of FedSoup is almost the same as that of FedRep in terms of previously unseen DG on the Retina dataset, which can be attributed to the very small number of clients (a mere two) within the FL framework. This limitation curtails the FedSoup's capability to accumulate diverse models, subsequently impacting its efficacy in achieving flat minima.

| Task | Metric | FedAvg | FedProx | MOON | FedBN | FedFomo | FedRep | FedBABU | **FedSoup** |
|---|---|---|---|---|---|---|---|---|---|
| Path. | Acc. | $68.72_{(1.05)}$ | $67.84_{(1.16)}$ | $70.47_{(1.19)}$ | $65.36_{(0.75)}$ | $60.32_{(0.63)}$ | $68.05_{(0.50)}$ | $70.92_{(1.04)}$ | $\mathbf{71.97}_{(1.19)}$ |
| | AUC | $76.76_{(1.03)}$ | $75.02_{(1.07)}$ | $78.08_{(1.10)}$ | $72.50_{(0.93)}$ | $64.28_{(0.74)}$ | $74.22_{(0.68)}$ | $78.80_{(1.07)}$ | $\mathbf{79.63}_{(1.10)}$ |
| Retina | Acc. | $63.96_{(3.28)}$ | $59.38_{(3.89)}$ | $63.90_{(4.27)}$ | $61.58_{(3.92)}$ | $61.53_{(2.65)}$ | $\underline{65.02_{(2.45)}}$ | $63.71_{(3.05)}$ | $64.32_{(1.27)}$ |
| | AUC | $67.39_{(2.31)}$ | $61.62_{(3.91)}$ | $66.03_{(2.93)}$ | $65.88_{(4.02)}$ | $64.11_{(3.37)}$ | $\underline{68.47_{(1.53)}}$ | $64.70_{(2.39)}$ | $\underline{68.24_{(1.02)}}$ |

**Table 3:** Local and global performance results comparison with SOTA PFL methods on the Pathology dataset (no hold-out client).

| Method | Local Performance | | Global Performance | |
| --- | --- | --- | --- | --- |
| | Accuracy ↑ | AUC ↑ | Accuracy ↑ | AUC ↑ |
| FedAvg | $86.52_{(0.85)}$ | $93.26_{(0.52)}$ | $79.65_{(0.99)}$ | $88.14_{(0.99)}$ |
| FedProx | $85.74_{(0.91)}$ | $93.28_{(0.45)}$ | $79.44_{(0.41)}$ | $87.89_{(0.64)}$ |
| MOON | $88.62_{(0.76)}$ | $95.06_{(0.18)}$ | $85.63_{(0.96)}$ | $93.20_{(0.51)}$ |
| FedBN | $88.48_{(0.15)}$ | $94.67_{(0.29)}$ | $77.31_{(1.37)}$ | $85.53_{(1.22)}$ |
| FedFomo | $83.40_{(3.39)}$ | $89.71_{(2.72)}$ | $65.92_{(1.29)}$ | $70.11_{(2.16)}$ |
| FedRep | $86.93_{(1.50)}$ | $93.15_{(1.41)}$ | $78.04_{(1.62)}$ | $85.09_{(1.42)}$ |
| FedBABU | $88.55_{(0.39)}$ | $94.92_{(0.44)}$ | $83.94_{(3.21)}$ | $91.48_{(2.67)}$ |
| **FedSoup** | $\mathbf{89.20}_{(0.53)}$ | $\mathbf{95.08}_{(0.61)}$ | $\mathbf{86.44}_{(1.12)}$ | $\mathbf{93.46}_{(0.74)}$ |



\end{document}

%% file: mathcommands.tex
\usepackage{amsmath,amsthm,amsfonts,bm}

\DeclareMathAlphabet{\mathsfit}{\encodingdefault}{\sfdefault}{m}{sl}
\SetMathAlphabet{\mathsfit}{bold}{\encodingdefault}{\sfdefault}{bx}{n}

\renewcommand{\hat}{\widehat}

%% file: sec_1_introduction.tex
\section{Introduction}

Federated learning (FL) has emerged as a promising methodology for harnessing the power of private medical data without necessitating centralized data governance~\cite{rieke2020future,dou2021federated,dayan2021federated,pati2022federated}. However, recent study~\cite{DBLP:journals/corr/abs-2206-09262} has identified a significant issue in current FL algorithms, namely, the trade-off between local and global performance when encountering distribution shifts. This issue is particularly prevalent in medical scenarios~\cite{li2020multi,DBLP:conf/iclr/LiJZKD21,jiang2022dynamic}, where medical images may undergo shifts of varying degrees due to differences in imaging device vendors, parameter settings, and the patient demographics. 
Personalized FL (PFL) techniques are typically utilized to address the  data heterogeneity problem by weighing more on in-distribution (ID) data of each client. For instance, %
FedRep~\cite{DBLP:conf/icml/CollinsHMS21} learns the entire network during local updates and keeps part of the network from global synchronization. However, they have a risk of overfitting to local data~\cite{DBLP:conf/icml/QuLDLTL22}, especially when client local data is limited, and have poor generalizability on out-of-distribution (OOD) data. Another line of work has studied the heterogeneity issue by regularizing the updates of local model. For instance, FedProx~\cite{DBLP:conf/mlsys/LiSZSTS20} constraints local updates to be closer to the global model.
An effective way to evaluate FL's generalizability is to investigate its performance on the joint global distribution following \cite{DBLP:journals/corr/abs-2206-09262}, which refers to testing the FL models on $\bigcup\{\mathcal{D}_i\}$, where $\mathcal{D}_i$ indicates client $i$'s distribution~\footnote{In the heterogeneous setting ($\mathcal{D}_i \neq \mathcal{D}_j$), $\mathcal{D}_j$ is viewed as the OOD data for client $i$.}.
Unfortunately, the existing works have not found the sweet spot between personalized (local) and consensus (global) models.

In this regard, we study a practical problem of enhancing personalization and generalization jointly in cross-silo FL for medical image classification when faced data heterogeneity.  To this end, we aim to address the following two questions in FL: \emph{What could be the causes that result in local and global trade-off?} and \emph{How to achieve better local and global trade-off?} First, we provide a new angle to understand the trade-off. We reveal that over-personalization in FL can cause overfitting on local data and trap the model into a sharp valley of loss landscape (highly sensitive to parameter perturbation, see detailed definition in Sec.~\ref{sec:trade}), thus limiting its generalizability. An effective strategy for avoiding sharp valleys in the loss landscape is to enforce models to obtain flat minima. In the centralized domain, weight interpolation has been explored as a means of seeking flat minima as its solution is moved closer to the centroid of the high-performing models, which corresponds to a flatter minimum~\cite{DBLP:conf/uai/IzmailovPGVW18,DBLP:conf/nips/ChaCLCPLP21,dayan2021federated,DBLP:journals/corr/abs-2212-10445}. However, research on these interpolation methods has been overlooked in FL.

With the above basis, we propose to track both local and global models during the federated training and perform model interpolation to seek the optimal balance. Our insight is drawn from the model soup method~\cite{DBLP:conf/icml/WortsmanIGRLMNF22}, which shows that averaging weights of multiple trained models with same initial parameters can enhance model generalization. 
However, the original model soup method requires training substantial models with varying hyper-parameters, which can be prohibitively time-consuming and costly in terms of communication during FL. Given the high communication cost and the inability to restart training from scratch in FL, we leverage global models at different time points within a single training session as the ingredients to adapt the model soup method~\cite{DBLP:conf/icml/WortsmanIGRLMNF22} to FL.

In this paper, we propose a novel federated model soup method (FedSoup) to produce an ensembled model from local and global models that achieve better local-global trade-off. We refer the `soup' as a combo of different federated models.
Our proposed FedSoup includes two key modules. The first one is temporal model selection, which aims to select suitable models to be combined into one.  The second module is Federated model patching~\cite{DBLP:journals/corr/abs-2208-05592}{, which refers to a 
fine-tuning technique that aims to enhance personalization without compromising the already satisfactory global performance.} 
For the first module, temporal model selection, we utilize a greedy model selection strategy based on the local validation performance. This avoids incorporating models that could be located in a different error landscape basin than the local loss landscape (shown in Figure \ref{fig:framework}). 
Consequently, each client possesses their personalized global model soups, consisting of a subset of historical global models that are selected based on their local validation sets. 
As for the second module, federated model patching, it introduces model patching in local client training by interpolating the local model and the global model soups into a new local model, 
bridging the gap between local and global domains. It promotes the personalization of the model for ID testing and also maintains good global performance for OOD generalization.

In summary, our key contributions are as follows: 
(i) A novel FL method called Federated Model Soups (FedSoup) is proposed to improve generalization and personalization by promoting smoothness and seeking flat minima. (ii) A new temporal model selection mechanism is designed for FL, which maintains a client-specific model soups with temporal history global model to meet personalization requirements while not incurring additional training costs. (iii) An innovative federated model patching method between local and global models is introduced in federated client training to alleviate overfitting of local limited data.

\input{figure/framework.tex}

%% file: figure/framework.tex
\begin{figure}[t]
    \centering
    \includegraphics[width=1.0\textwidth]{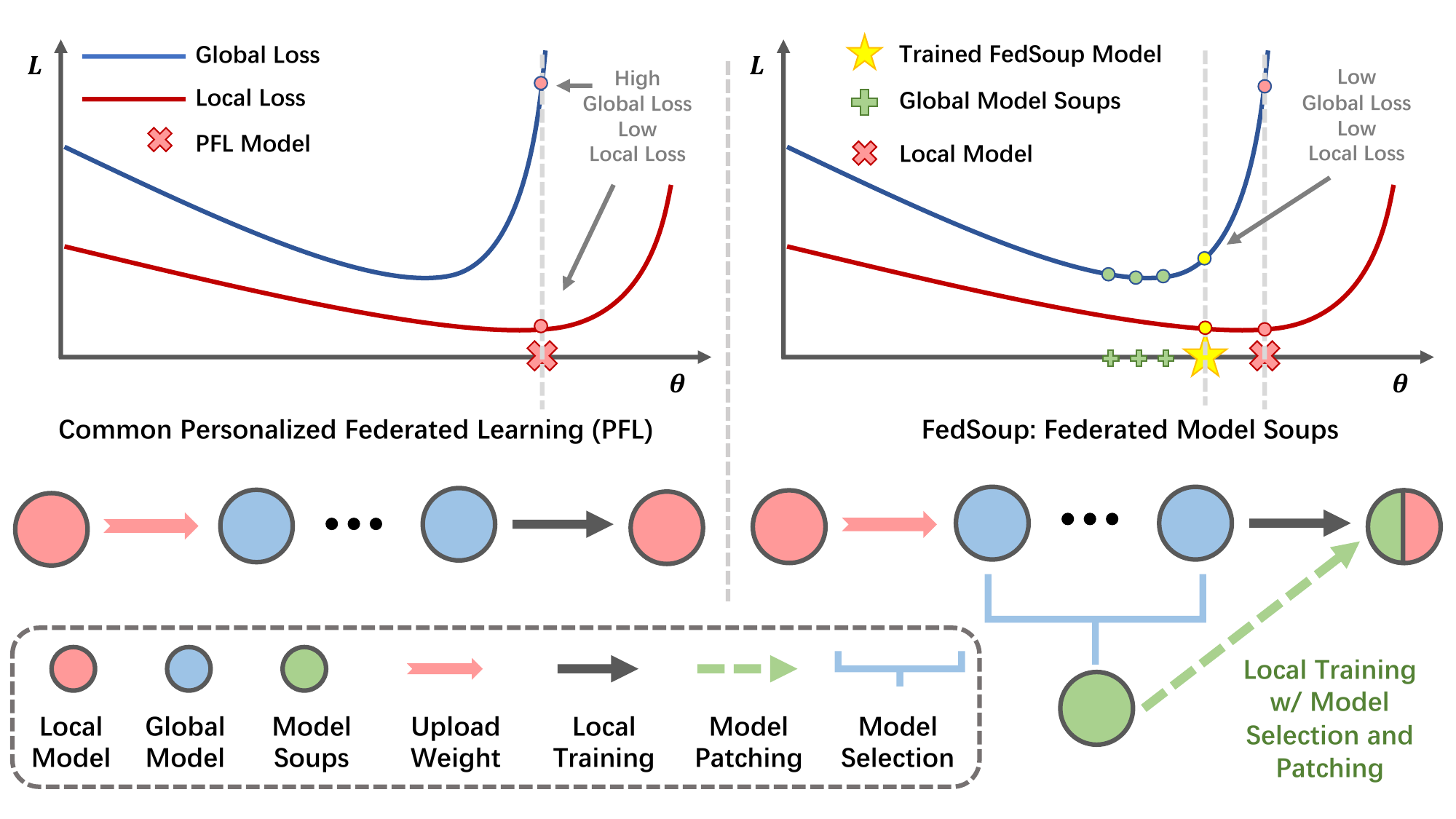} 
    \vspace{-0mm}
    \caption{The overview of our method (FedSoup) compared with common PFL methods. PFL methods typically minimize local loss but suffering high global loss. While our federated model soup method balances both local and global loss by seeking flat minima. The black dots in the figure represent ellipsis and indicate multiple rounds of model upload and model training in between. Compared to previous pFL methods, our approach introduces global model selection modules and local model interpolation with the global model (referred to as model patching).}
    \label{fig:framework}
    \vspace{-0mm}
\end{figure}

%% file: sec_2_method.tex
\section{Method}

\subsection{Problem Setup}

Consider a cross-silo FL setting with $N$ clients. Let $\mathcal{D}:=\left\{\mathcal{D}_{i}\right\}_{i=1}^{N}$ be a set of $N$ training domain, each of which is a distribution over the input space $\mathcal{X}$.
For each client, we have access to $n$ training data points in the form of $(x_{j}^{i},y_{j}^{i})_{j=1}^{n}\sim\mathcal{D}_{i}$, where $y_{j}^{i}$ denotes the target label for input $x_{j}^{i}$. 
We also define a set of unseen target domains $\mathcal{T}:=\left\{\mathcal{T}_{i}\right\}_{i}^{N'}$ in a similar manner, where $N'$ is the number of target domains and is typically set to one.
The goal of personalization (local performance) is to find a model $f(\cdot;\theta)$ via minimizing an empirical risk $\hat{\mathcal{E}}_{\mathcal{D}{i}}(\theta):=\frac{1}{n}\sum_{j=1}^{n}\ell(f(x^{i};\theta), y^{i}))$ over a local client training set $\mathcal{D}_i$, where $\ell(\cdot,\cdot)$ is a loss function. 
On the other hands, the objective of generalization (global performance) is to minimize both population loss $\mathcal{E}_{\mathcal{D}}(\theta)$ and $\mathcal{E}_{\mathcal{T}}(\theta)$ over multiple domains by Empirical Risk Minimization (ERM) $\hat{\mathcal{E}}_{\mathcal{D}}(\theta):=\frac{1}{Nn}\sum_{i=1}^{N}\sum_{j=1}^{n}\ell(f(x^{i};\theta), y^{i}))$ over all training clients' training sets $\mathcal D$. 
In this work, we evaluate the local performance on local testing samples from $\mathcal{D}_i$, and evaluate the global performance on testing samples from the joint global distribution $\mathcal{D}:=\left\{\mathcal{D}_{i}\right\}_{i=1}^{N}$.

\subsection{Generalization and Flat Minima}
\label{sec:trade}
In practice, ERM in deep neural networks, \textit{i.e.}, $\arg\min_\theta \hat{\mathcal{E}}_{\mathcal{D}}(\theta)$, can yield multiple solutions that offer comparable training loss, but vastly different levels of generalizability~\cite{DBLP:conf/nips/ChaCLCPLP21}. 
However, without proper regularization, models are prone to overfit the training data and the training model will fall into a sharp valley of the loss surface, which is less generalizable~\cite{DBLP:conf/iclr/ChaudhariCSLBBC17}.

One common reason for failures in ERM is the presence of variations in the data distribution ($\mathcal{D}_{i} \neq \mathcal{D}$), which can cause a shift in the loss landscape. 
As illustrated in Figure \ref{fig:framework}, {the sharper the optimized minima, the more sensitive it is to shifts in the loss landscape. This results in an increase in generalization error. 
In cross-silo FL, each client may overfit their local training data, leading to poor global performance. This is due to the distribution shift problem, which creates conflicting objectives among the local models \cite{DBLP:conf/icml/QuLDLTL22}. 
Therefore, when the local model converges to a sharp minima, the higher the degree of personalization (local performance) of the model, the more likely it is to have poor generalization ability (global performance)}.

From the domain generalization formalization in~\cite{DBLP:conf/nips/ChaCLCPLP21,DBLP:journals/ml/Ben-DavidBCKPV10}, the test loss $\mathcal{E}_{\mathcal{T}}(\theta)$ can be bounded by the robust empirical loss $\hat{\mathcal{E}}_{\mathcal{D}}^{\epsilon}(\theta)$ as follows: 
\begin{equation}
\vspace{-1.5mm}
\label{eq:dg_bound}
\mathcal{E}_{\mathcal{T}}(\theta) < \hat{\mathcal{E}}_{\mathcal{D}}^{\epsilon}(\theta) +\frac{1}{N}\sum\nolimits_{i=1}^{N}\sup_{A \in \mathcal{A}}|\mathcal{P}_{\mathcal{D}_{i}}(A)-\mathcal{P}_{\mathcal{T}}(A)|+\xi,
\end{equation}
where the $\sup_{A}|\mathcal{P}_{\mathcal{D}_{i}}(A)-\mathcal{P}_{\mathcal{T}}(A)|$ is a divergence between domain $\mathcal{D}_{i}$ and $\mathcal{T}$, $\mathcal{A}$ is the set of measurable subsets under $\mathcal{D}_{i}$ and $\mathcal{T}$, and $\xi$ is the confidence bound term related to the radius $\epsilon$ and the number of the training samples.
From the Equation~\eqref{eq:dg_bound}, we can infer that minimizing sharpness and seeking flat minima is directly related with the generalization performance on the unseen target domain.

\subsection{Our Solution: FedSoup}

After analyzing the aforementioned relationship between sharpness and generalization, we expound on the distinctive challenges of seeking flat minima and mitigating the trade-off between local and global performance in FL. Consequently, we introduce two refined modules as the ingredient of our proposed FedSoup solution. FedSoup only needs to modify the training method of the client, and the algorithm implementation is shown in Algorithm \ref{alg:fedsoup}.

\noindent\textbf{Temporal Model Selection.}
Stochastic Weight Averaging (SWA) \cite{DBLP:conf/uai/IzmailovPGVW18} and Sharpness-Aware Minimization (SAM) \cite{DBLP:conf/iclr/ForetKMN21} are two commonly used flat minima optimizers, which seek to find parameters in wide low-loss basins. 
In contrast to SAM, which incurs extra computational cost to identify the worst parameter perturbation, SWA is a more succinct and effective approach for implicitly favoring the flat minima by averaging weights. 
The SWA algorithm is motivated by the observation that SGD often finds high-performing models in the weight space but rarely reaches the central points of the optimal set. By averaging the parameter values over iterations, the SWA algorithm moves the solution closer to the centroid of this space of points.

Nevertheless, when it comes to cross-silo FL training, the discrepancy between clients is significant, and models might lie in different basins. Merging all these models haphazardly is not effective and might hinder generalization. 
Recently, a selective weight averaging method called model soups~\cite{DBLP:conf/icml/WortsmanIGRLMNF22} was introduced to enhance the generalization of fine-tuned models.
The original model soups is not applicable to the FL setting, requiring high communication costs and training compute. 
We adapt the idea to a new approach by leveraging global models trained at different time points in one pass of FL training. 
Additionally, considering the heterogeneity of data distribution in cross-silo FL and the requirement of personalization, we propose a model selection strategy where each client utilizes the performance of its local validation set as a monitoring indicator.
We called this module temporal model selection (see Algorithm \ref{alg:fedsoup} Line $7\text{-}8$)

\noindent\textbf{Federated Model Patching.}
According to previous analysis on the loss landscape, there is a loss landscape offset between different FL clients due to their domain discrepancy. Thus, simply integrating a global model can damage the model's personalization. To address this, we introduce the use of the model patching \cite{DBLP:journals/corr/abs-2208-05592} (\textit{i.e.} local and global model interpolation) during client-side local training in FL, aiming to improve model personalization and maintain the good global performance.
Specifically, 
model patching approach forces local client not to distort global model severely and seek low-loss interpolated model between local and global, encouraging the local and global model lie in the same basin without a large barrier of linear connectivity.  \cite{DBLP:conf/iclr/MirzadehFGP021}. We called this module federated model patching. 
In summary, the update rule of FedSoup is implemented as follows:
\begin{equation}
\label{eq:swa}
\theta_{FedSoup} \gets \frac{\theta_g^1 + \dots + \theta_g^k + \theta_l}{k + 1},
\end{equation}
where $\theta_g$ is global model, $\theta_l$ is local model, $k$ is the number of selected global models. This update rule corresponds to Algorithm \ref{alg:fedsoup} Line $9$.

\input{algorithm/fedsoup_algorithm.tex}

It is important to note that our proposed FedSoup algorithm requires only one carefully tuned hyper-parameter, namely the interpolation start epoch. To mitigate the risk of having empty global model soups when the start epoch is too late and to prevent potential performance degradation when the start epoch is too early, we have set the default interpolation start epoch to be 75\% of the total training epochs, aligning with the default setting of SWA. 
Furthermore, it is worth mentioning that the modified model soup and model patching modules in our proposed FedSoup framework are interdependent. Model patching, which is a technique based on our modified model soup algorithm, provides an abundance of models to explore flatter minima and enhance performance.

%% file: algorithm/fedsoup_algorithm.tex
\begin{algorithm}[tb]
\caption{$\mathsf{FedSoup}$}
\label{alg:fedsoup}
    \begin{algorithmic}[1]
    \STATE {\bfseries Input:} global model $\theta_{g}$, local model $\theta_{l}$, last epoch local model $\hat{\theta}^{i}_l$, number of clients $k$, number of epochs $n$, interpolation start epoch $E$.
    \STATE $ \mathsf{soup} \gets \{ \}$
    \STATE {\bfseries for} $i = 0$ {\bfseries to} $n$ {\bfseries do}
    \STATE \ \ \ $\theta_g \gets \mathsf{Aggregation}( \hat{\theta}^{1}_l, \dots, \hat{\theta}^{k}_l )$
    \STATE \ \ \ $\theta_l \gets \mathsf{ClientUpdate}( \theta_g )$
    \STATE \ \ \ {\bfseries if} $i \geq E$ {\bfseries then}
    \STATE \ \ \ \ \ \ {\bfseries if}  $\mathsf{ValAcc}\mleft(\mathsf{average}\mleft(\mathsf{soup} \cup \{ \theta_l \}\cup \{ \theta_g \}\mright)\mright)\geq \mathsf{ValAcc}\mleft(\mathsf{average}\mleft(\mathsf{soup} \cup \{ \theta_l \}\mright)\mright)$ {\bfseries then}
    \STATE \ \ \ \ \ \ \ \ \ \ $\mathsf{soup} \gets \{\theta_g \}$ \hfill\COMMENT{Module 1: Temporal Model Selection}
    \STATE \ \ \ \ \ \ $\theta_l \gets \mathsf{average}\mleft(\mathsf{soup} \cup \{ \theta_l \}\mright)$ \hfill\COMMENT{Module 2: Federated Model Patching}
    \STATE {\bfseries return} $\theta_l$
    \end{algorithmic}
\end{algorithm}

%% file: sec_3_experiments.tex
\section{Experiments}

\subsection{Experimental Setup}

\subsubsection{Dataset.}
We validate the effectiveness of our proposed method, FedSoup, on two medical image classification tasks. The first task involved the classification of pathology images 
from five different sources using Camelyon17 dataset\cite{DBLP:journals/tmi/BandiGMDBHBLPZL19}, and each source is viewed as a client. The pathology experiment consists of a total of $4,600$ images \footnote{We take a random subset from the original Camelyon17 dataset to match the small data settings in FL\cite{DBLP:conf/aistats/McMahanMRHA17}.}, each with a resolution of $96 \times 96$. The second task involved retinal fundus images from four different institutions \cite{DBLP:conf/cbms/FumeroASSG11,sivaswamy2015comprehensive,DBLP:journals/mia/OrlandoFBKBDFHK20}, and each institute is viewed as a client. The retinal fundus experiment consists of a total of $1,264$ images, each with a resolution of $128 \times 128$. The objective of both datasets is to identify abnormal images from normal ones. We also maintained an equal number of samples from each client to prevent clients with more data from having a disproportionate influence on the global performance evaluation.

\noindent\textbf{Evaluation Setup.}
For each client, we take  $75\%$ of the data as the training set. To assess the generalization ability and personalization of our model, we have constructed both local and global testing sets. Following~\cite{DBLP:journals/corr/abs-2206-09262}, in our experimental setting, we first create a held-out global testing set by randomly sampling an equal number of images per source/institute, so its distribution is different from either of the client. 
The local testing dataset for each FL client is the remaining sample from the same source as its training set. The number of local testing set per client is approximately the same as that of the held-out global testing set. For the pathology dataset, as each subject can have multiple samples, we have ensured that data from the same subject only appeared in either the training or testing set. To align with the cross-validation setting for subsequent out-of-domain evaluations, we conducted a five-fold leave-one-client-data cross-validation, with three repetitions using different random seeds in each fold. The results of the repeated experiments without hold-out client data are provided in the appendix. 
For PFL methods, we report the performance by averaging the results of each personalized models.

\noindent\textbf{Models and Training Hyper-Paramters.}
We employ the ResNet-18 architecture as the backbone model. Our approach initiates local-global interpolation at the $75\%$ training phase, consistent with the default hyper-parameter setting of SWA. We utilize the Adam optimizer with learning rate of $1\mathrm{e}{-3}$, momentum coefficients of $0.9$ and $0.99$ and set the batch size to $16$. We set the local training epoch to 1 and perform a total of $1,000$ communication rounds.

\subsection{Comparison with State-of-the-art Methods}

\input{table/main_result.tex}

We compare our method with seven common FL and state-of-the-art PFL methods. 
Results in Table~\ref{table:main} demonstrate that our FedSoup method achieves competitive performance on local evaluation sets while significantly improving generalization with respect to global performance.
Furthermore, FedSoup exhibits greater stability with lower variance of performance across multiple experiments.

Comparing the performance improvement of FedSoup across different datasets, we observed that FedSoup had a more substantial effect on the smaller retinal fundus dataset compared to the larger pathology dataset. 
In terms of performance gap compared to the second-best methods (FedBABU on Retina and FedProx on Pathology), our approach demonstrates a larger advantage on the Retina dataset.
This observation suggests that our proposed method can mitigate the negative impact of local overfitting caused by small local datasets, thus improving the model's generalization ability.

\noindent\textbf{Sharpness Quantification.}
The sharpness measure used in this study calculates the median of the dominating Hessian eigenvalue across all training set batches using the Power Iteration algorithm~\cite{DBLP:conf/bigdataconf/YaoGKM20}. This metric indicates the maximum curvature of the loss landscape, which is often used in the literature on flat minima \cite{DBLP:journals/corr/abs-2202-00661} to reflect the sharpness. The median of the dominating Hessian eigenvalue of all clients in the retinal fundus dataset was measured in this part. 
Based on the Figure \ref{fig:sub:sharpness} presented, it is evident that the proposed method leads to flatter minima as compared to the other methods. 

\noindent\textbf{Trade-off at Different Personalized Levels.}
Following the evaluation in \cite{DBLP:journals/corr/abs-2206-09262}, we conducted an experiment comparing the local and globalperformance of different models at various levels of personalization using the retinal fundus datasets. We control the personalization level for the PFL methods by varying the number of iterations that the model undergoes fine-tuning using only the local training set after federated training. As we increase the number of fine-tuning iterations, we consider the level of personalization to be higher. We choose to show model performance after fine-tuning iterations $1$, $7$, and $15$.
The results in Figure \ref{fig:sub:per_level} indicated that existing FL methods often have a trade-off between local and global performance. As the number of fine-tuning iterations increases, local performance typically improves but global performance decreases. 
Compared to other methods, our appraoch maintains high local performance while also preventing a significant drop in global performance, which remains much higher than other methods.

\input{figure/exp_fig.tex}

\subsection{Unseen Domain Generalization}
We show the additional benefits of FedSoup on unseen domain generalization.
\noindent\textbf{Setup.}
To evaluate the generalization of our method beyond the participating domains, we utilize one domain that did not take part in the distributed training and used its data as the evaluation set for unseen domain generalization. 
To this end, we perform leave-one-out cross-validation by having one client as the to-be-evaluated unseen set each time. To ensure a reliable results of unseen domain generalization, we conducted experiments on the Camelyon17 dataset, which has a larger number of samples.

\noindent\textbf{Results.}
Overall, our proposed method demonstrates an advantage in terms of unseen domain generalization capabilities (see Figure \ref{fig:sub:oof}). In comparison to FedAvg, Our approach resulted in a 2.87-point increase in the AUC index for generalization to unseen domains on the pathology dataset.

%% file: table/main_result.tex
\begin{table}[t]
    \centering
    \caption{Local and global performance results comparison with SOTA PFL methods on two medical image classification tasks.}
    \vspace{-3mm}
    \label{table:main}
    \setlength{\aboverulesep}{0.5pt}
    \setlength{\belowrulesep}{0.5pt}
    \resizebox{1.0\columnwidth}{!}{
    \begin{tabular}{l*8c}
        \toprule
        & \multicolumn{4}{c}{Pathology} & \multicolumn{4}{c}{Retinal Fundus} \\
        \cmidrule(lr){2-5} \cmidrule(lr){6-9}
        Method & \multicolumn{2}{c}{Local Performance} & \multicolumn{2}{c}{Global Performance} & \multicolumn{2}{c}{Local Performance} & \multicolumn{2}{c}{Global Performance} \\
        \cmidrule(lr){2-3} \cmidrule(lr){4-5} \cmidrule(lr){6-7} \cmidrule(lr){8-9}
        & Accuracy $\uparrow$  & AUC $\uparrow$ & Accuracy $\uparrow$ & AUC $\uparrow$ & Accuracy $\uparrow$ & AUC $\uparrow$ & Accuracy $\uparrow$ & AUC $\uparrow$ \\    
        \midrule
        FedAvg \cite{DBLP:conf/aistats/McMahanMRHA17} & $82.41_{(0.61)}$ & $90.44_{(0.42)}$ & $70.18_{(1.25)}$ & $78.74_{(1.31)}$ & $89.81_{(1.17)}$ & $95.24_{(0.73)}$ & $73.27_{(3.97)}$ & $81.02_{(4.57)}$ \\ 
        FedProx \cite{DBLP:conf/mlsys/LiSZSTS20} & $\textbf{86.34}_{(0.52)}$ & $\textbf{92.78}_{(0.36)}$ & $67.42_{(1.32)}$ & $77.18_{(1.32)}$ & $90.14_{(0.20)}$ & $95.47_{(0.63)}$ & $70.46_{(2.36)}$ & $76.84_{(2.93)}$ \\ 
        MOON \cite{DBLP:conf/cvpr/LiHS21} & $85.71_{(0.42)}$ & $91.98_{(0.34)}$ & $70.61_{(1.54)}$ & $79.27_{(1.43)}$ & $89.14_{(0.77)}$ & $94.91_{(0.98)}$ & $77.49_{(3.15)}$ & $84.89_{(2.70)}$ \\ 
        FedBN \cite{DBLP:conf/iclr/LiJZKD21} & $82.32_{(0.87)}$ & $90.07_{(0.77)}$ & $65.16_{(0.69)}$ & $71.61_{(0.97)}$ & $89.70_{(1.45)}$ & $95.49_{(0.56)}$ & $75.11_{(0.52)}$ & $82.70_{(1.01)}$ \\ 
        FedFomo \cite{DBLP:conf/iclr/ZhangSFYA21} & $80.99_{(1.07)}$ & $86.51_{(0.99)}$ & $61.00_{(0.59)}$ & $61.69_{(0.78)}$ & $89.70_{(0.00)}$ & $94.88_{(0.45)}$ & $59.90_{(2.00)}$ & $63.82_{(1.14)}$ \\ 
        FedRep \cite{DBLP:conf/icml/CollinsHMS21} & $82.50_{(0.53)}$ & $89.77_{(0.43)}$ & $66.87_{(0.60)}$ & $72.34_{(0.75)}$ & $89.38_{(0.89)}$ & $95.16_{(0.83)}$ & $71.81_{(1.79)}$ & $79.09_{(2.44)}$ \\ 
        FedBABU \cite{DBLP:journals/corr/abs-2106-06042} & $85.18_{(0.33)}$ & $92.39_{(0.29)}$ & $69.56_{(1.40)}$ & $77.26_{(1.48)}$ & $90.25_{(1.39)}$ & $95.10_{(0.33)}$ & $77.65_{(0.24)}$ & $85.09_{(0.21)}$ \\ 
        \textbf{FedSoup} & $85.71_{(0.37)}$ & $92.47_{(0.31)}$ & $\textbf{72.87}_{(1.35)}$ & $\textbf{81.45}_{(1.40)}$ & $\textbf{90.92}_{(0.50)}$ & $\textbf{96.00}_{(0.43)}$ & $\textbf{78.64}_{(0.90)}$ & $\textbf{86.24}_{(0.86)}$ \\ 
        \bottomrule
    \end{tabular}}
\end{table}

%% file: figure/exp_fig.tex
\begin{figure}[t]
    \centering
    \setlength{\abovecaptionskip}{-0.8mm}
    \subfigure[] {
        \label{fig:sub:sharpness}
        \includegraphics[width=0.305\columnwidth]{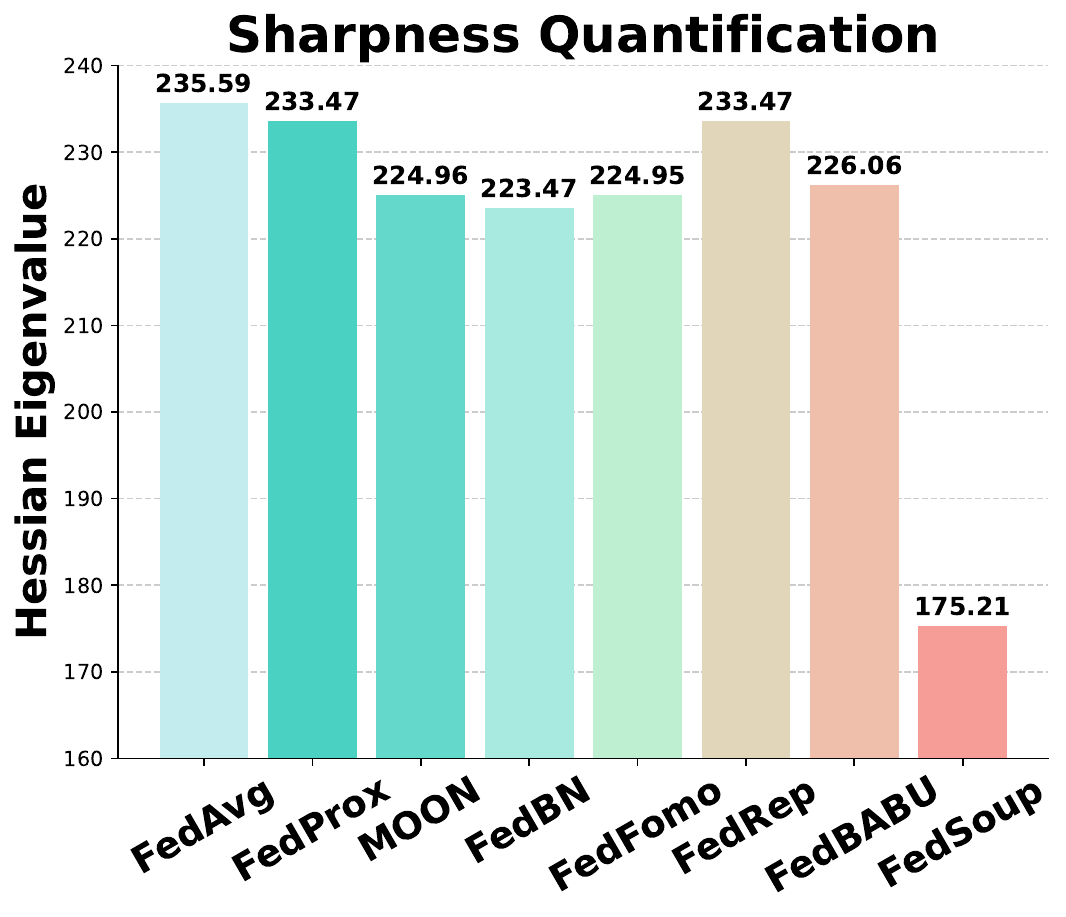}
    }
    \subfigure[] {
        \label{fig:sub:per_level}
        \includegraphics[width=0.32\columnwidth]{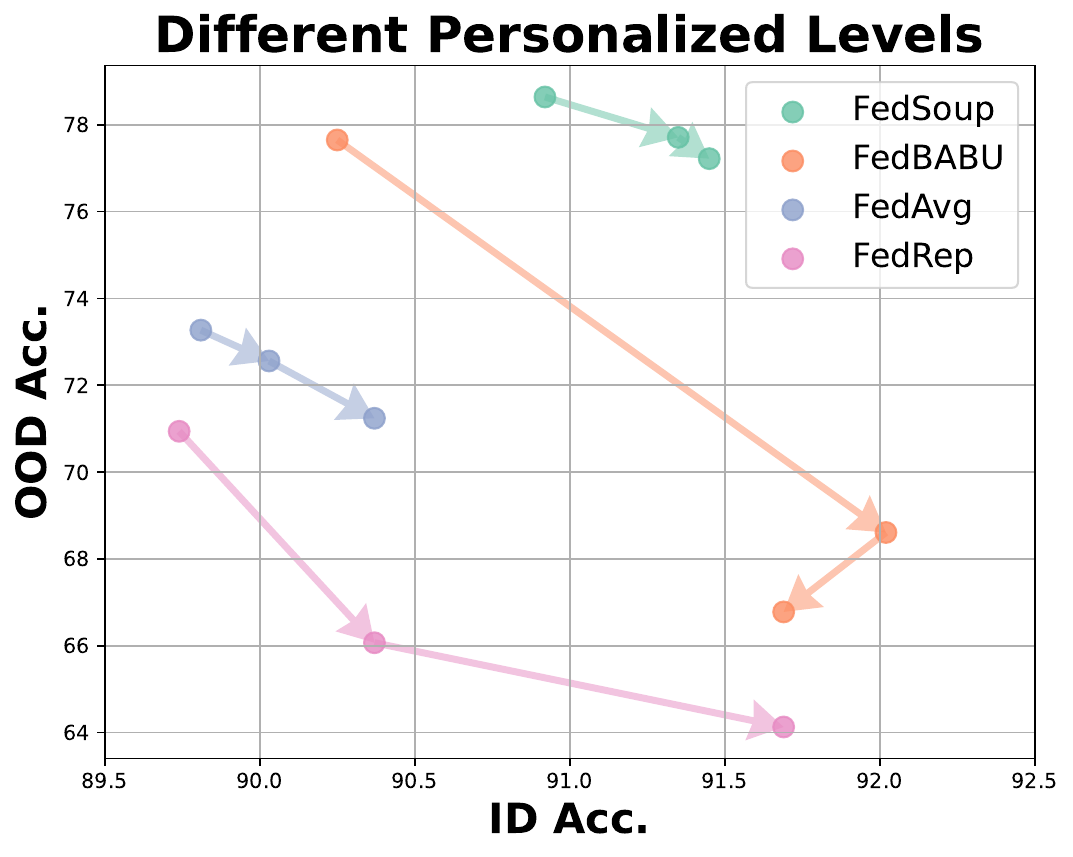}
    }
    \subfigure[] {
        \label{fig:sub:oof}
        \includegraphics[width=0.305\columnwidth]{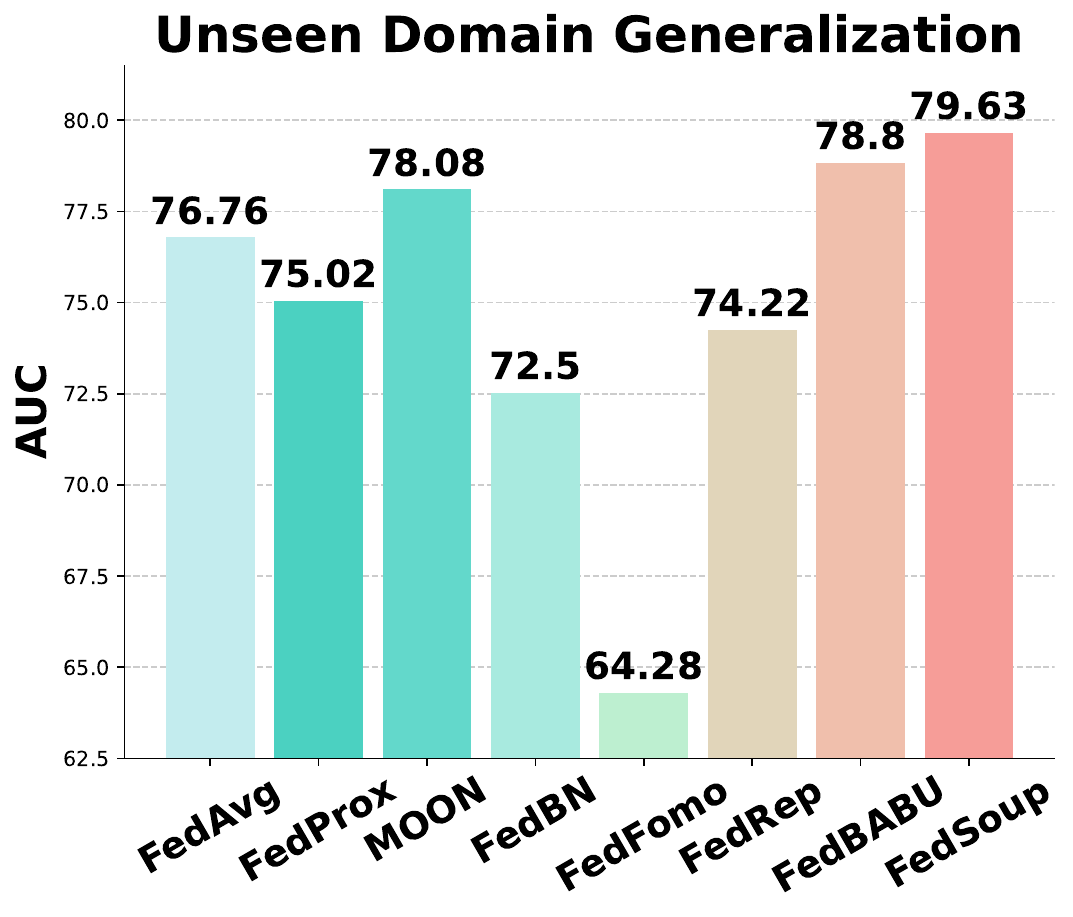}
    }
    \caption{Analysis of our approach: (a) sharpness quantification on the retina dataset, (b) local and global trade-off under different personalized levels (fine-tuning epochs), (c) unseen domain generalization on the pathology dataset.}
    \label{fig:exp_fig}
    \vspace{-4mm}
\end{figure}

%% file: sec_4_conclusion.tex
\section{Conclusion}

In this paper, we demonstrate the trade-off between personalization and generalization in the current FL methods for medical image classification. To optimize this trade-off and achieve flat minima, we propose the novel FedSoup method. By maintaining personalized global model pools in each client and interpolating weights between local and global models, our proposed method enhances both generalization and personalization. FedSoup outperforms other PFL methods in terms of both generalization and personalization, without incurring any additional inference or memory costs.